\documentclass[runningheads]{llncs}

\usepackage{eccv}

\usepackage{eccvabbrv}

\usepackage{graphicx}
\usepackage{booktabs}
\usepackage{amsmath}
\usepackage{amssymb}
\usepackage{multirow}
\usepackage{algorithm}
\usepackage{algorithmic}

\usepackage[accsupp]{axessibility}  

\usepackage[pagebackref,breaklinks,colorlinks,citecolor=eccvblue]{hyperref}

\usepackage{orcidlink}

\begin{document}

\title{Viewpoint Matters: Dynamically Optimizing Viewpoints with Masked Autoencoder for Visual Manipulation} 

\titlerunning{Viewpoint Matters}

\author{Pengfei Yi\inst{1,2} \and Yifan Han\inst{1,2} \and Junyan Li\inst{1,2}
\and Litao Liu\inst{3} \and Wenzhao Lian\inst{4*}}

\authorrunning{Pengfei Yi et al.}

\institute{Institute of Automation, Chinese Academy of Sciences \and
the School of Artificial Intelligence, University of Chinese Academy of Sciences \and Rutgers University \and School of Artificial Intelligence, Shanghai Jiao Tong University}

\maketitle

\begin{abstract}
  Robotic manipulation continues to be a challenge, and imitation learning (IL) enables robots to learn tasks from expert demonstrations. Current IL methods typically rely on fixed camera setups, where cameras are manually positioned in static locations, imposing significant limitations on adaptability and coverage. Inspired by human active perception, where humans dynamically adjust their viewpoint to capture the most relevant and least noisy information, we propose MAE-Select, a novel framework for active viewpoint selection in single-camera robotic systems. MAE-Select fully leverages pre-trained multi-view masked autoencoder representations and dynamically selects the next most informative viewpoint at each time chunk without requiring labeled viewpoints. Extensive experiments demonstrate that MAE-Select improves the capabilities of single-camera systems and, in some cases, even surpasses multi-camera setups. The project will be available at \href{https://sites.google.com/view/mae-select}{https://sites.google.com/view/mae-select}.
  \keywords{Manipulation \and Imitation Learning \and Active Perception}
\end{abstract}

\section{Introduction}
\label{sec:intro}

\begin{figure}[t]
\begin{center}
\includegraphics[width=\textwidth]{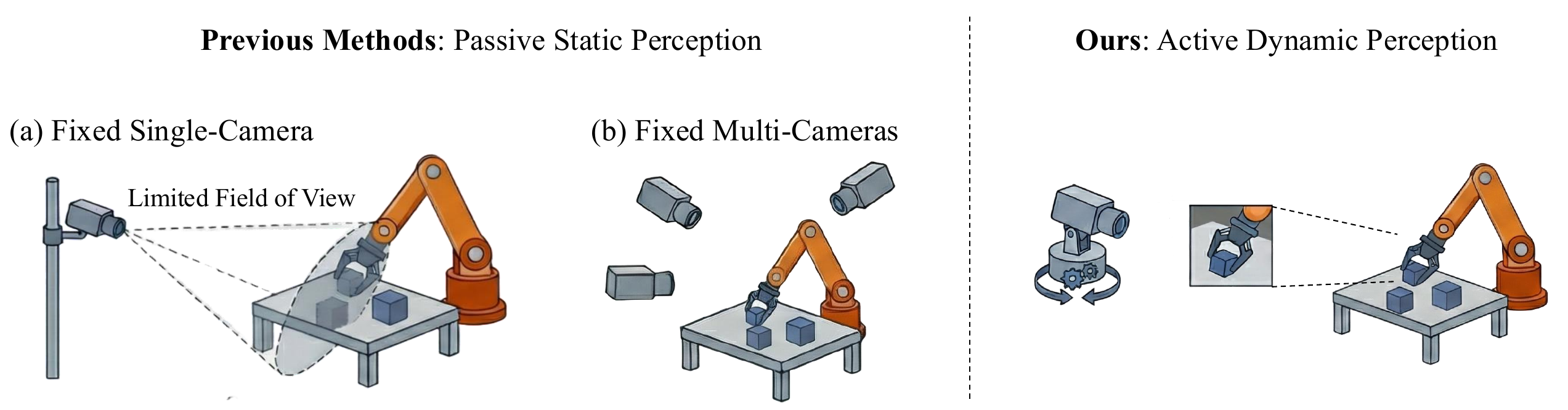}
\end{center}
   \caption{Comparison of visual perception paradigms for robotic manipulation. Traditional passive static setups (left), such as fixed single-camera or multi-camera systems, often suffer from visual occlusions and data redundancy. In contrast, our proposed MAE-Select framework (right) employs active dynamic perception to efficiently acquire optimal task-relevant viewpoints.}
\label{fig:abstract}
\end{figure}

Robotic manipulation is a core challenge in robotics, critical to applications ranging from industrial automation to healthcare. Imitation Learning (IL) \cite{pomerleau1988alvinn,jang2022bc,florence2022implicit,ebert2021bridge,brohan2022rt,liu2024foam} has become a leading approach for enabling robots to learn complex tasks through expert demonstrations. Recently, advances in deep generative models \cite{goodfellow2020generative,huang2023diffusion}, such as variational autoencoders (VAEs) \cite{kingma2013auto} and diffusion models \cite{ho2020denoising}, have empowered IL by allowing robots to process high-dimensional sensory inputs, such as images, leading to promising results in robotic manipulation \cite{chi2023diffusion,zhao2023learning,bharadhwaj2024roboagent,reuss2023goal}. 

However, most current IL methods rely on fixed camera setups, either single or multiple cameras, which pose significant limitations. In fixed single-camera setups \cite{pari2021surprising,qin2022one}, though practical and cost-effective, robots face challenges due to the limited field of view, which may obstruct critical parts of the environment or objects, negatively impacting task performance. Multi-camera setups, while designed to provide more comprehensive scene coverage, introduce their own complexities: the abundance of redundant or irrelevant information can overwhelm learning algorithms and decrease efficiency. As shown in Sec.\ref{sec:experiments}, these passive static multi-view setups do not always provide the most task-relevant and cleanest information, leading to inefficiencies and suboptimal decision-making.

In contrast, humans dynamically adjust their viewpoints while performing tasks. By actively moving our head, we naturally seek the most informative, least noisy perspectives. Inspired by this human capability, we propose shifting from passive, static perception to \textbf{active perception}, where the viewpoint is dynamically adjusted throughout the task to optimize information intake. In a practical robotic setting, this could be embodied by a robot moving its head to capture the most task-relevant views in real-time. In this paper, we focus on the feasibility of active viewpoint selection for robotic manipulation as an initial exploration along this direction.

To this end, we introduce \textbf{MAE-Select}, a framework designed to actively select optimal viewpoints for single-camera robotic setups. MAE-Select first fully utilizes the powerful pre-trained representations from the multi-view masked autoencoders (MAEs) \cite{he2022masked, tong2022videomae}, leveraging its complete encoder-decoder architecture to obtain estimated multi-view representations. Unlike prior works that focus on fixed viewpoints \cite{seo2023multi}, MAE-Select dynamically predicts the next better viewpoint based on the current chunk of visual and action information. Crucially, this viewpoint selection is learned solely through imitation learning, which does not require manual labels for optimal views. It demonstrates significant potential in the advancement of single-camera robotic manipulation.

Our key contributions are as follows:

\begin{itemize}
    \item We propose MAE-Select, a novel mechanism that dynamically selects the next optimal viewpoint at each time chunk without manual labels.
    \item  We present an IL framework that fully exploits pre-trained MAE representations for manipulation.
    \item We demonstrate through various experiments that MAE-Select enhances manipulation accuracy in single-camera setups, even outperforming multi-camera systems in certain tasks.
\end{itemize}

\section{Related Work}
\subsection{Imitation Learning for Manipulation}

Imitation learning (IL), which enables agents to acquire complex skills by observing and mimicking expert behavior, has become a cornerstone approach for robotic manipulation \cite{avigal2022speedfolding,arunachalam2023holo,de2019causal,duan2017one,jang2022bc,johns2021coarse,wen2022catgrasp,li2025splatter,han2026fsag}. A classic method within this paradigm is behavioral cloning (BC) \cite{pomerleau1988alvinn}, which formulates the problem as supervised learning by directly mapping observations to actions. Despite its simplicity, BC remains widely used due to its ease of implementation and scalability.

Recent years have seen a surge of IL techniques leveraging high-capacity neural networks and large-scale visual observations, driven by advances in deep learning and data collection platforms. Several methods now incorporate high-dimensional visual modalities such as RGB images \cite{florence2019self,rahmatizadeh2018vision,wen2022you,zhao2023learning,chi2023diffusion}, 3D point clouds \cite{ze20243d,chen2021unsupervised}, and virtual reality demonstrations \cite{zhang2018deep}, allowing policies to learn rich perceptual representations. For instance, ACT \cite{zhao2023learning} introduces a transformer-based action chunking strategy that improves sample efficiency and generalization. Meanwhile, Diffusion Policy \cite{chi2023diffusion} formulates policy learning as a conditional denoising process, achieving stable and expressive visuomotor behavior across a range of complex tasks.

Despite these advances, most IL frameworks operate under fixed-camera configurations, limiting their adaptability to spatial variations and occlusions. While recent works have investigated multi-view learning or bird’s-eye views, these approaches often require multiple calibrated cameras or rely on extensive 3D reconstruction pipelines \cite{shridhar2023perceiver}. In contrast, our work emphasizes the importance of selecting informative viewpoints from a single camera, seeking to bridge the gap between rich multi-view information and practical deployment constraints in manipulation tasks.

\subsection{Active Perception and Next Best View Planning}

The concept of active perception, famously pioneered by Bajcsy~\cite{bajcsy1988active}, posits that an agent should dynamically alter its sensor state to improve perceptual understanding. In robotics, this is often formulated as the Next Best View (NBV) problem, where the goal is to determine the optimal sensor placement for tasks such as 3D reconstruction, object recognition, or navigation~\cite{connolly1985determination, delmerico2018comparison}. Traditional NBV approaches rely on information-theoretic metrics, such as minimizing entropy or maximizing expected information gain based on volumetric representations like octrees~\cite{potthast2014probabilistic}. 

Recently, with the rise of deep reinforcement learning (RL) and neural rendering, active vision has been applied to dynamic environments. Works utilizing Neural Radiance Fields (NeRF)~\cite{mildenhall2021nerf} or 3D Gaussian Splatting~\cite{kerbl20233d} have explored active camera trajectories to minimize rendering uncertainty~\cite{pan2022activenerf}. Furthermore, RL has been employed to learn camera control policies that maximize task-specific rewards~\cite{zhu2025active}. However, these methods often require complex reward engineering or extensive 3D reconstruction pipelines. In contrast, MAE-Select implicitly learns the optimal viewpoint selection through an imitation learning objective, directly tying active perception to the downstream manipulation task without requiring explicit NBV heuristics or ground-truth viewpoint labels.

\subsection{Unsupervised Representation Learning}

Unsupervised representation learning aims to extract semantically meaningful features from unlabeled data, enabling downstream tasks to benefit from improved generalization and data efficiency \cite{sohn2015learning,jing2020self,donahue2016adversarial}. Among the leading approaches, Masked Autoencoders (MAE) \cite{he2022masked} have emerged as a powerful technique, training models to reconstruct masked portions of the input by learning context-aware latent representations. This pretext task has led to significant advances in vision applications, including image classification, segmentation, and video understanding \cite{li2022uniform,zhang2022hivit,tong2022videomae}.

In the context of robotics, several works have incorporated MAE-based frameworks to improve policy learning from visual inputs. R3M \cite{nair2022r3m} and Real-world MAE \cite{radosavovic2023real} show that unsupervised pretraining can provide robust and generalizable representations for manipulation. Specifically, MV-MWM \cite{seo2023multi} extends MAE to multi-view settings by encoding multiple perspectives during training and applying single-view inference at test time, demonstrating enhanced performance on several manipulation benchmarks. However, its viewpoint remains static, using predefined camera without exploring alternative viewpoints. In contrast, our work explicitly investigates the viewpoint selection problem, aiming to discover the most informative camera view for manipulation tasks.

\begin{figure*}
\begin{center}
\includegraphics[width=\textwidth]{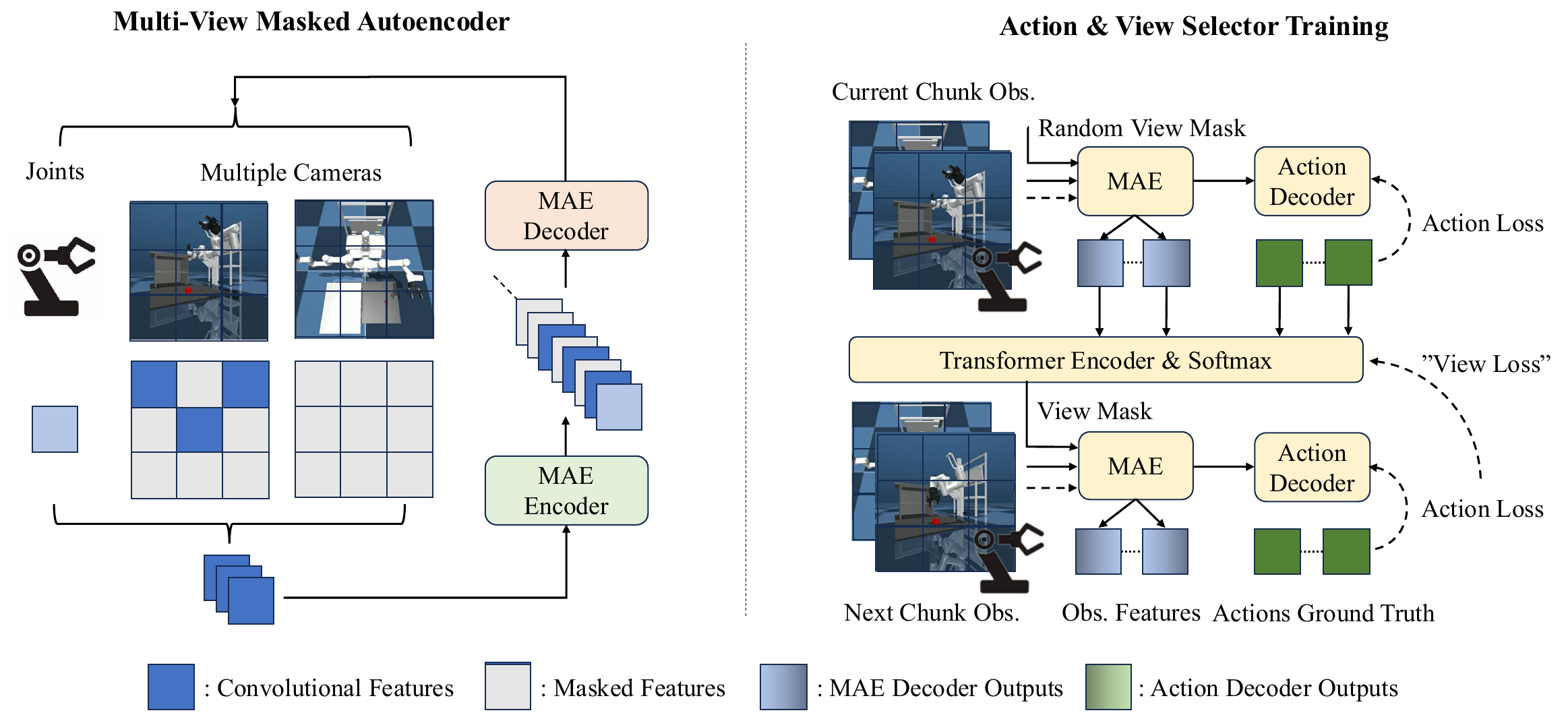}
\end{center}
   \caption{Illustration of our proposed method. \textbf{Left} depicts the pre-training stage of the multi-view masked autoencoder with joint embeddings. \textbf{Right} illustrates the training process of our framework using imitation learning.}
\label{fig:pipeline}
\end{figure*}

\section{Method}
\label{sec:method}

We introduce \textbf{MAE-Select}, a novel framework that enables robotic agents to actively select informative viewpoints for complex manipulation tasks. The core of our approach is a new formulation in which optimal viewpoint selection is learned implicitly through an imitation learning objective—without requiring explicit supervision or reinforcement learning. In contrast to prior work that relies solely on the encoder of pre-trained models \cite{seo2023masked, seo2023multi}, our method leverages the full generative capacity of a multi-view masked autoencoder (MV-MAE), including both its encoder and decoder. This enables the agent to build a rich, 3D-aware representation of the scene from just a single viewpoint, which is essential for accurate action prediction and informed viewpoint selection. An overview of our architecture is shown in Fig.\ref{fig:pipeline}. We begin by formalizing the problem, then describe our representation learning strategy, and finally explain how the action and viewpoint selection policies are trained jointly.

\subsection{Problem Formulation}

We frame the robotic manipulation task as a learning problem over a sequence of observations and actions. Let the agent's state at time $t$ be composed of its proprioceptive information $s_t$ (e.g., joint angles) and a set of visual observations $O_t = \{o_t^v\}_{v \in V}$ from a fixed set of $N_v$ available viewpoints $V$. The agent's goal is to learn a policy $\pi$ that takes the current state and generates a chunk of $T$ future actions, $a_{t:t+T-1} = \{a_t, \dots, a_{t+T-1}\}$.

In our active perception setting, the policy does not have access to all viewpoints simultaneously. Instead, it operates on a single selected viewpoint $o_t^{v_t}$ at time $t$. Therefore, the objective is to learn a composite policy $\pi$ that can be decomposed into two key components:
\begin{enumerate}
    \item \textbf{An action policy $\pi_{\theta}(a_{t:t+T-1} | o_t^{v_t}, s_t)$}: This policy predicts a sequence of future actions based on the current single-view observation and proprioceptive state.
    \item \textbf{A view selection policy $\pi_{\psi}(v_{t+T} | o_t^{v_t}, s_t, a_{t:t+T-1})$}: This policy selects the most advantageous viewpoint, $v_{t+T}$, for the subsequent time chunk, conditioned on the information from the current chunk.
\end{enumerate}
Both policies, parameterized by $\theta$ and $\psi$ respectively, are trained jointly from a dataset of expert demonstrations $\mathcal{D} = \{ (O_t, s_t, a_t) \}_{t=0 \dots T_m}^N$ using an imitation learning framework.

\subsection{Multi-View Masked Autoencoder Pre-training}
To build a strong foundation for visual understanding, we pre-train a Multi-View Masked Autoencoder (MV-MAE) on the demonstration data. This pre-training stage, depicted in the left panel of the pipeline diagram, is designed to learn a compressed yet comprehensive representation of the multi-view scene.

The architecture is trained to reconstruct a full multi-view image set from a heavily masked input. Given a multi-view observation $O_t$, we first extract convolutional feature maps $\{\bar{o}_t^v \in \mathbb{R}^{H \times W \times D_{feat}}\}_{v \in V}$. We then apply a dual-masking strategy:
\begin{itemize}
    \item \textbf{Patch Masking}: A large percentage of the feature patches within each view's map are randomly masked (i.e., zeroed out).
    \item \textbf{View Masking}: Entire views are randomly masked, forcing the model to infer inter-view relationships.
\end{itemize}
The set of unmasked feature patches, augmented with learned 1D view-specific embeddings and fixed 2D sinusoidal position embeddings, are fed into a Transformer-based encoder, $f_{\phi}$, to produce a latent representation $z_t^m$. The decoder, $g_{\phi}$, then takes this latent code, a full set of mask tokens, and the robot's joint state embedding $s_t$ to reconstruct the original, unmasked features for all viewpoints. The pre-training objective is to minimize the reconstruction error:
\begin{equation}
\mathcal{L}_{\text{MAE}}(\phi) = \mathbb{E}_{O_t, s_t \sim \mathcal{D}} \left[ \| O_t - g_{\phi}(f_{\phi}(O_t^m), s_t) \|^2_2 \right]
\end{equation}
where $O_t^m$ denotes the masked input views. This process endows the model with a powerful generative prior, enabling it to hallucinate a full 3D scene representation from a single, potentially occluded input viewpoint during the downstream task.

\subsection{Next Better Viewpoint Selection}
\label{subsec:viewpoint_rewritten}
Following pre-training, we fine-tune the model and train the action and viewpoint selection policies concurrently, as illustrated in the middle panel of the diagram. Our central contribution is a training strategy for the viewpoint selector $\pi_\psi$ without explicit labels for the "best" viewpoint. We achieve this by using the action prediction loss of a future time chunk as a supervisory signal for the viewpoint choice made in the current chunk. This process unfolds over two consecutive chunks of data from a trajectory, $\mathcal{D}_t$ and $\mathcal{D}_{t+T}$.

\begin{algorithm}[t]
\caption{Training MAE-Select (Single Iteration)}
\label{alg:mae_select}
\begin{algorithmic}[1]
\REQUIRE Expert demonstration dataset $\mathcal{D}$, Pre-trained MV-MAE encoder $f_\phi$ and decoder $g_\phi$, Action policy $\pi_\theta$, Viewpoint selector $\pi_\psi$, Chunk size $T$.
\STATE Sample consecutive data chunks $(\mathcal{D}_t, \mathcal{D}_{t+T})$ from $\mathcal{D}$.
\STATE \textbf{// Step 1: Process Current Chunk}
\STATE Sample an initial viewpoint $v_t \sim \mathcal{U}(V)$.
\STATE Extract single-view observation $o_t^{v_t}$ from $\mathcal{D}_t$.
\STATE Compute multi-view context: $C_t = g_\phi(f_\phi(o_t^{v_t}), s_t)$.
\STATE Sample noise $\epsilon_t$ and timestep $k \sim \mathcal{U}(1, K)$.
\STATE Compute action loss: $\mathcal{L}_{\text{action}}^{(t)} = \| \epsilon_t - \pi_\theta(C_t, a_{t:t+T-1}^0 + \epsilon_t, k) \|^2$.
\STATE \textbf{// Step 2: Select Next Viewpoint via STE}
\STATE Predict viewpoint probabilities: $c_{\text{hat}} = \text{Softmax}(\pi_\psi(C_t, a_{t:t+T-1}))$.
\STATE Discretize to one-hot vector: $\mathbf{y}_{\text{hard}} = \text{argmax}(c_{\text{hat}})$.
\STATE Apply Straight-Through Estimator: $\mathbf{y} = c_{\text{hat}} + \text{sg}[\mathbf{y}_{\text{hard}} - c_{\text{hat}}]$.
\STATE \textbf{// Step 3: Process Next Chunk}
\STATE Select next observation using $\mathbf{y}$: $o_{t+T}^{\hat{v}} = \mathbf{y} \cdot O_{t+T}$.
\STATE Compute new context: $C_{t+T} = g_\phi(f_\phi(o_{t+T}^{\hat{v}}), s_{t+T})$.
\STATE Sample new noise $\epsilon_{t+T}$ and timestep $k'$.
\STATE Compute next action loss: $\mathcal{L}_{\text{action}}^{(t+T)} = \| \epsilon_{t+T} - \pi_\theta(C_{t+T}, a_{t+T:t+2T-1}^0 + \epsilon_{t+T}, k') \|^2$.
\STATE \textbf{// Step 4: Optimization}
\STATE Compute total loss: $\mathcal{L}_{\text{total}} = \mathcal{L}_{\text{action}}^{(t)} + \lambda_{1} \mathcal{L}_{\text{action}}^{(t+T)} + \lambda_{2} \mathcal{L}_{\text{MAE}}$.
\STATE Update $\theta, \phi$ using $\nabla \mathcal{L}_{\text{total}}$.
\STATE Update $\psi$ using gradients flowing back from $\mathcal{L}_{\text{action}}^{(t+T)}$ via STE.
\end{algorithmic}
\end{algorithm}

\textbf{Processing the Current Chunk ($\mathcal{D}_t$):}
For the current chunk starting at time $t$, we begin with a single view $o_t^v$, selected uniformly at random. This single-view observation is passed through the entire pre-trained MV-MAE ($f_\phi, g_\phi$) to generate an estimated multi-view feature context $C_t = g_\phi(f_\phi(o_t^v), s_t)$. This context $C_t$ is then fed into the action decoder $\pi_\theta$—a diffusion-based policy—which is trained to predict the noise $\epsilon_t$ added to the expert action trajectory $a_{t:t+T-1}$. The action loss for this chunk is computed as:
\begin{equation}
\mathcal{L}_{\text{action}}^{(t)} = \mathbb{E}_{\epsilon_t, k} \left[ \| \epsilon_t - \pi_\theta(C_t, a_{t:t+T-1}^0 + \epsilon_t, k) \|^2 \right]
\end{equation}
where $k \sim \mathcal{U}(1, K)$ is a randomly sampled diffusion timestep.

\textbf{Selecting the View for the Next Chunk ($\mathcal{D}_{t+T}$):}
The viewpoint selector $\pi_\psi$, implemented as a Transformer encoder, takes the feature context $C_t$ and the ground-truth action trajectory $a_{t:t+T-1}$ from the current chunk to predict a probability distribution over the available viewpoints for the next chunk:
\begin{equation}
c_{\text{hat}} = \text{Softmax}(\pi_\psi(C_t, a_{t:t+T-1}))
\end{equation}
To select a discrete viewpoint while maintaining differentiability for backpropagation, we employ a straight-through estimator (STE) like VQ-VAE \cite{van2017neural}. In the forward pass, we perform an argmax operation to get a one-hot vector $\mathbf{y}_{\text{hard}}$ representing the chosen view. For the backward pass, gradients flow through the continuous softmax probabilities $c_{\text{hat}}$. This is implemented as:
\begin{equation}
\mathbf{y} = c_{\text{hat}} + sg[\mathbf{y}_{\text{hard}} - c_{\text{hat}}]
\end{equation}
where $\text{sg}[\cdot]$ denotes the stop-gradient operator. This mechanism enables the gradient from the subsequent action loss to inform the selection policy.

\textbf{Processing the Next Chunk ($\mathcal{D}_{t+T}$):}
The one-hot vector $\mathbf{y}$ is used to select a single viewpoint $o_{t+T}^{\hat{v}}$ from the next observation set $O_{t+T}$. This selection is performed via a dot product $o_{t+T}^{\hat{v}} = \mathbf{y} \cdot O_{t+T}$, which also enables gradient flow back to the selection policy. This selected view is then processed through the identical pipeline as the first chunk, yielding a new feature context $C_{t+T}$ and a corresponding action loss, $\mathcal{L}_{\text{action}}^{(t+T)}$.

\textbf{Updating the Viewpoint Selector:}
There is no explicit "View Loss" term. Instead, the training signal for the viewpoint selector $\pi_\psi$ is implicitly provided by the action loss of the subsequent time chunk, $\mathcal{L}_{\text{action}}^{(t+T)}$. Through the use of the STE, gradients from $\mathcal{L}_{\text{action}}^{(t+T)}$ are backpropagated through the action decoder and the selected observation to update the parameters $\psi$ of the selection model. This process directly optimizes the selector to choose viewpoints that minimize future action prediction error, a mechanism visually represented by the "View Loss" feedback loop in the training diagram.

The comprehensive training objective, therefore, combines the losses from both time chunks with the auxiliary reconstruction objective to train all components end-to-end:
\begin{equation}
\mathcal{L}_{\text{total}} = \mathcal{L}_{\text{action}}^{(t)} + \lambda_{1} \mathcal{L}_{\text{action}}^{(t+T)} + \lambda_{2} \mathcal{L}_{\text{MAE}}
\end{equation}
By optimizing this single objective, the action policy $\pi_\theta$ is updated by both action loss terms, while the viewpoint selector $\pi_\psi$ is specifically optimized by the gradients flowing from the future action loss, $\mathcal{L}_{\text{action}}^{(t+T)}$.

\begin{figure}
\begin{center}
\includegraphics[width=\textwidth]{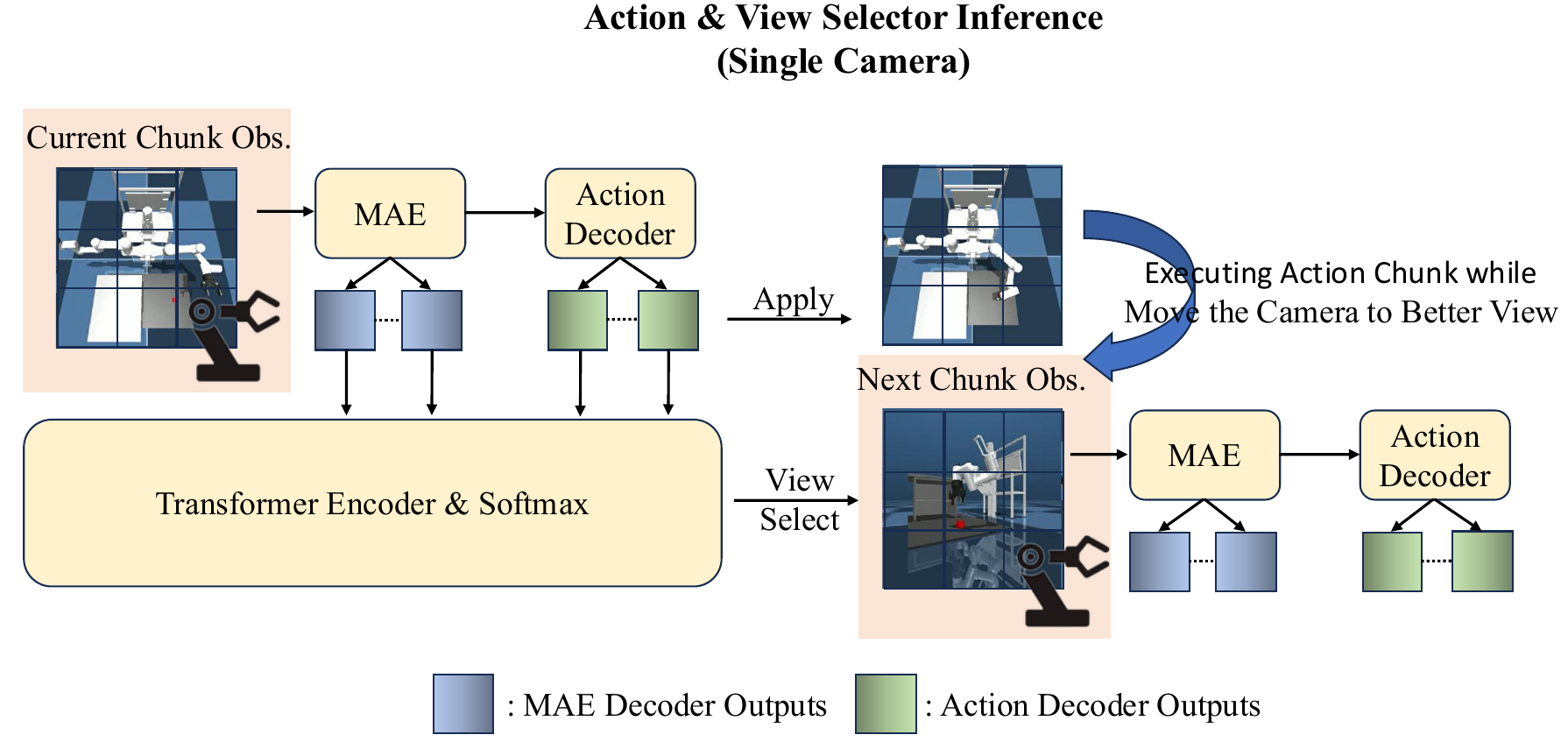}
\end{center}
   \caption{Overview of the inference pipeline for the proposed framework.}
\label{fig:pipeline_infer}
\end{figure}

\textbf{Inference:} During inference, as shown in Fig.\ref{fig:pipeline_infer}, the process is autoregressive. Starting with a random initial viewpoint, the agent uses the observation to predict both the first action chunk and the optimal viewpoint for the next chunk. The predicted action is then used alongside the newly selected viewpoint for the subsequent step, creating a dynamic perception-action loop.

\begin{figure}
\begin{center}
\includegraphics[width=0.95\textwidth]{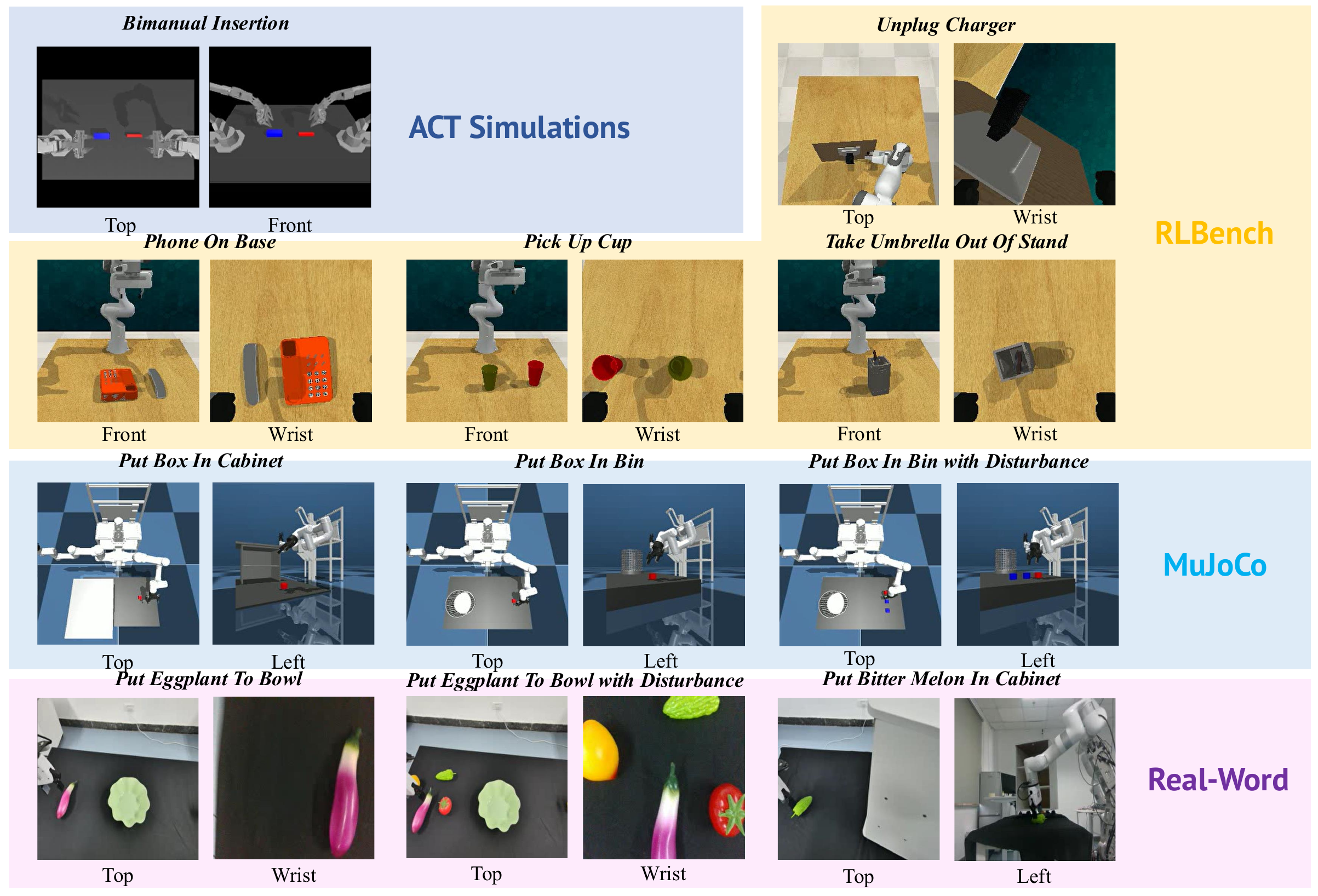}
\end{center}
   \caption{The viewpoint settings for various robotic tasks, showcasing the viewpoints used to evaluate performance across different simulation and real-world scenarios.}
\label{fig:tasks}
\end{figure}

\section{Experiments}
\label{sec:experiments}

We evaluate MAE-Select across 3 challenging scenarios and 8 demanding tasks in simulation, including simulations in ACT \cite{zhao2023learning}, RLBench \cite{james2019rlbench}, and our designed robot in MuJoCo \cite{todorov2012mujoco}, along with three real-world tasks.

\subsection{Tasks}
The 11 tasks cover various scenarios, providing a comprehensive evaluation of our method. The viewpoint settings are shown in Fig.~\ref{fig:tasks}. We categorize these tasks across four different environments:

\subsubsection{ACT Simulations \cite{zhao2023learning}:} 
We focus on the \textit{Bimanual Insertion} task, where the robot's arms pick up a socket and a peg for a mid-air insertion.

\subsubsection{RLBench \cite{james2019rlbench}:} 
We select four varied tasks as follows:
\begin{itemize}
    \item \textit{Phone On Base}: The robot needs to pick up and place a phone onto its base.
    \item \textit{Pick Up Cup}: The robot needs to grasp and lift a specific cup from among several cups.
    \item \textit{Unplug Charger}: The robot is tasked with removing a charger from a socket, involving careful manipulation to avoid damaging delicate components.
    \item \textit{Take Umbrella Out Of Stand}: This task requires the robot to pick up a small umbrella.
\end{itemize}

\subsubsection{MuJoCo \cite{todorov2012mujoco}:} 
We design three customized tasks for our robot:
\begin{itemize}
    \item \textit{Put Box In Cabinet}: The robot needs to pick up a box and place it inside a cabinet, requiring precise spatial reasoning.
    \item \textit{Put Box In Bin}: In this task, the robot places a box into a bin, testing its ability to interact with constrained spaces.
    \item \textit{Put Box In Bin with Disturbance}: Similar to the previous task, but the robot must select a specific block from multiple blocks.
\end{itemize}

\subsubsection{Real-World Evaluation:} 
We designed three tasks similar to the simulation setups but involving real objects: \textit{Put Bitter Melon In Cabinet}, \textit{Put Eggplant To Bowl}, and \textit{Put Eggplant To Bowl with Disturbance}.

\subsection{Data Collection}

Following previous methods \cite{zhao2023learning,james2019rlbench}, we
collect demonstrations with a scripted policy for each task in simulation. The datasets capture the robot’s proprioceptive data, action sequences, and multi-view visual observations. For real-world data acquisition, we implement a low-cost teleoperation platform inspired by Gello \cite{wu2024gello}. Our robot consists of a 7DOF Ufactory xarm 7 robotic arm and a parallel-jaw gripper. Our camera setup includes two statically mounted (top and left) and one wrist-mounted Realsense D435 camera. For each task, we collected 50 expert demonstrations, while the initial placement of the objects is varied uniformly in 2D regions. The final dataset comprises RGB data from two cameras with a resolution of 224×224, along with joint states and action sequences.

\subsection{Implementation}

We explore a single-camera control setup where the system is trained with multiple camera views but deployed with a single moving camera. At each time chunk, the camera is placed at one of the training viewpoints, enabling flexible operation in real-world settings where only one camera is available. We base our implementation on the architecture of the diffusion policy \cite{chi2023diffusion}. The action space corresponds to the joint angles of the robot arm, while the image observations have a resolution of $224\times 224$ with a patch size of 16. Our masked autoencoder utilizes a 12-layer ViT \cite{dosovitskiy2020image} for encoder and an 8-layer ViT for decoder, with an embedding dimension of 512. During pretraining, we use a batch size of 128 over 100 epochs. For RLBench and real-world tasks, the time chunk size is set to 20, with a total of 600 epochs. For other tasks, the time chunk size is 100, with 1,000 epochs in total. $\lambda_{1}$ and $\lambda_{2}$ are empirically set to 2.0 and 1.0, respectively. The batch size is set to 32, and the training undergoes two stages, where only the imitation component is trained during the first half of the epochs, and both imitation and view selection components are trained in the second half. 
For each method, we evaluate the best-performing checkpoints from the last three evaluated at 100-intervals, with 50 environment initializations in simulation. All models were trained and tested on NVIDIA RTX 4090 GPUs.

\begin{table}
\caption{Results of comparison experiment. * represents with Disturbance. For the eight simulated tasks, we report the success rate, with 50 policy evaluations each. For the real-world tasks, we report the number of successes, training with human data, with 10 evaluations. For Diffusion Policy and MAE-Diffusion, the viewpoint data used for training is the same as the viewpoint used for testing. For MAE-Select, we use both viewpoints for training and only one viewpoint is used per time chunk during testing. Bold and underlined fonts mean the best and second-best results.}
\centering
\resizebox{\textwidth}{!}{
\begin{tabular}{ccccccccccccc}
\toprule
\multirow{2}{*}{Method} & \multicolumn{3}{c}{Bimanual Insertion} & \multicolumn{3}{c}{Put Box In Cabinet} & \multicolumn{3}{c}{Put Box In Bin}  & \multicolumn{3}{c}{Put Box In Bin*}\\  
\cmidrule(lr){2-4}\cmidrule(lr){5-7}\cmidrule(lr){8-10}\cmidrule(lr){11-13}
                        & Top  & Front  & Both & Top  & Left & Both & Top  & Left & Both & Top & Left & Both \\ \midrule
    Diffusion Policy \cite{chi2023diffusion} &42\% &44\% & 50\% & 16\% & 18\%& 26\% & 80\% & 64\% & 84\% & 38\% & 30\% & 44\% \\
    MAE-Diffusion &48\% & 50\%& \textbf{54\%}& 42\% & 42\%& \underline{46\%} & 84\% & 78\% & \textbf{92\%} & 52\% & 46\% & \textbf{60\%} \\
    MAE-Select &\multicolumn{3}{c}{\underline{52\%}} & \multicolumn{3}{c}{\textbf{50\%}} & \multicolumn{3}{c}{\underline{88\%}} & \multicolumn{3}{c}{\underline{58\%}} \\
\midrule
\multirow{2}{*}{Method} & \multicolumn{3}{c}{Phone On Base} & \multicolumn{3}{c}{Pick Up Cup} & \multicolumn{3}{c}{Take Umbrella}  & \multicolumn{3}{c}{Unplug Charger}\\  
\cmidrule(lr){2-4}\cmidrule(lr){5-7}\cmidrule(lr){8-10}\cmidrule(lr){11-13}
                        & Front  & Wrist  & Both & Front  & Wrist & Both & Front  & Wrist & Both & Top & Wrist & Both \\ \midrule
    Diffusion Policy \cite{chi2023diffusion} &82\% & 56\% & 78\% & 60\% & 40\%& 64\% & 58\% & 36\% & 54\% & 44\% & 30\% & 34\% \\
    MAE-Diffusion &86\% & 70\% & \underline{88\%} & \underline{68\%} & 66\%& 62\% & 56\% & 42\% & \textbf{64\%} & 46\% & 34\% & \underline{52\%} \\
    MAE-Select &\multicolumn{3}{c}{\textbf{92\%}} & \multicolumn{3}{c}{\textbf{70\%}} & \multicolumn{3}{c}{\underline{60\%}} & \multicolumn{3}{c}{\textbf{58\%}} \\
\midrule
\multirow{2}{*}{Method} & \multicolumn{4}{c}{Put Eggplant To Bowl} & \multicolumn{4}{c}{Put Eggplant To Bowl*} & \multicolumn{4}{c}{Put Bitter Melon In Cabinet} \\  
\cmidrule(lr){2-5}\cmidrule(lr){6-9}\cmidrule(lr){10-13}
                        & Top  & \multicolumn{2}{c}{Wrist}& Both & Top  & \multicolumn{2}{c}{Wrist}& Both & Top  & \multicolumn{2}{c}{Left}& Both \\ \midrule
    Diffusion Policy \cite{chi2023diffusion} &2/10 & \multicolumn{2}{c}{1/10} & 5/10 & 2/10& \multicolumn{2}{c}{0/10} & 3/10 & 0/10 & \multicolumn{2}{c}{1/10} & 2/10 \\
    MAE-Diffusion &4/10 & \multicolumn{2}{c}{4/10} & \textbf{7/10} & \underline{4/10}& \multicolumn{2}{c}{3/10} & \textbf{6/10} & 2/10 & \multicolumn{2}{c}{\underline{4/10}} & \underline{4/10} \\
    MAE-Select &\multicolumn{4}{c}{\underline{6/10}} & \multicolumn{4}{c}{\textbf{6/10}} & \multicolumn{4}{c}{\textbf{5/10}} \\
\bottomrule
\end{tabular}
}
\label{tab:ep1}
\end{table}

\subsection{Results}



We compare the performance of MAE-Select with two other methods: Diffusion Policy \cite{chi2023diffusion} and a variant of Diffusion Policy that incorporates MAE, referred to as MAE-Diffusion, across different types of tasks and viewpoints. For Diffusion Policy and MAE-Diffusion, the observation viewpoint configuration is held consistent between training and testing phases. In contrast, MAE-Select employs a multi-view training strategy that incorporates both available viewpoints, while adopting a dynamic viewpoint selection mechanism during testing where only a single optimal viewpoint is activated per time chunk.

As demonstrated in Tab.\ref{tab:ep1}, MAE-Select consistently outperforms other fixed single-camera setups in both simulation and real-world experiments. For example, in the \textit{Put Box In Cabinet} task, MAE-Select improves performance by 8\% compared to the best fixed single-camera method and by 32\% compared to previous work. Its advantage lies in its ability to intelligently select the most informative viewpoints, which allows the system to make the most of limited visual input, resulting in optimized task completion.  

Furthermore, an interesting pattern emerges in some tasks: for certain methods, the performance with a single viewpoint can surpass that of a multi-camera setup. For example, in the \textit{Unplug Charger} task with Diffusion Policy, using only the top view (44\%) outperforms using both views (34\%). This counterintuitive result may stem from the added complexity of processing multiple cameras, which can introduce noise or misalignment issues, complicating the learning process. By focusing on the optimal viewpoint, MAE-Select avoids these challenges, enabling more efficient and effective task execution. Consequently, MAE-Select remains highly competitive when compared to multi-camera setups, even outperforming them in several tasks.

\subsection{Ablation Study}
We conduct a comprehensive series of ablation studies to validate the efficacy and necessity of the core components within our proposed MAE-Select architecture. This section primarily investigates the framework along two dimensions: compatibility with various action decoders and the utilization efficiency of the Masked Autoencoder (MAE) structure.

\begin{table}
\caption{Results of ACT experiments. We report the success rate, with 50 policy evaluations each. Bold and underlined fonts mean the best and second-best results.}
\centering
\begin{tabular}{ccccccccccccc}
\toprule
\multirow{2}{*}{Method} & \multicolumn{3}{c}{Bimanual Insertion} & \multicolumn{3}{c}{Phone On Base}\\  
\cmidrule(lr){2-4}\cmidrule(lr){5-7}
                        & Top  & Front  & Both & Front  & Wrist & Both \\ \midrule
    ACT &14\% & 26\% & 34\% & 56\% & 50\%& 58\%  \\
    MAE-ACT &28\% & 30\% & \textbf{42\%} & 60\% & 58\%& \underline{66\%}  \\
    MAE-Select &\multicolumn{3}{c}{\underline{36\%}} & \multicolumn{3}{c}{\textbf{70\%}} \\
\bottomrule
\end{tabular}
\label{tab:ep2}
\end{table}

\subsubsection{Action Decoder Compatibility}
A primary advantage of our method is the high decoupling of its viewpoint selection mechanism from the specific downstream action decoder. This implies our viewpoint selection module can serve as a perception component, flexibly adapting to various action generation models. To highlight this versatility, we also evaluate our approach in combination with ACT \cite{zhao2023learning}, an alternative action decoder. As shown in Table \ref{tab:ep2}, integrating our method significantly enhances baseline architectures. Notably, in the \textit{Phone On Base} task, MAE-Select reaches a 70\% success rate, outperforming both ACT (58\%) and MAE-ACT (66\%). This demonstrates that dynamic single-view selection can integrate seamlessly into various architectures, providing a sharper decision basis and lower input redundancy than fixed multi-view fusion.

\begin{table}
\centering
\caption{Ablation Studies on MAE Encoder and Decoder Utilization.}
\begin{tabular}{ccccccccccccc}
\toprule
\multirow{2}{*}{Method} & \multicolumn{3}{c}{Put Box In Cabinet} & \multicolumn{3}{c}{Phone On Base}\\  
\cmidrule(lr){2-4}\cmidrule(lr){5-7}
                        & Top  & Left  & Both & Front  & Wrist & Both \\ \midrule
    MAE-Encoder &20\% & 28\% & 34\% & 76\% & 56\%& 80\%  \\
    MAE-Diffusion &\textbf{42\%} & \textbf{42\%}& \textbf{46\%} & \textbf{86\%} & \textbf{70\%}& \textbf{88\%}  \\
\bottomrule
\end{tabular}
\label{tab:ep3}
\end{table}

\subsubsection{Masked Autoencoder}
We also conduct an ablation study to isolate the contributions of fully utilizing the entire encoder-decoder structure of the masked autoencoder in our approach. Specifically, we compare the performance of our method, which leverages both the encoder and decoder components, against a version that uses only the encoder of the masked autoencoder \cite{seo2023masked,seo2023multi}. The results in Tab.\ref{tab:ep3} demonstrate that utilizing the full encoder-decoder structure significantly improves performance, particularly in scenarios that require nuanced visual understanding from partial or occluded viewpoints. This highlights the value of the decoder in refining the system's ability to interpret and act based on incomplete or masked information, contributing to better generalization and adaptability in both simulated and real-world tasks.

\begin{figure}
\begin{center}
\includegraphics[width=\textwidth]{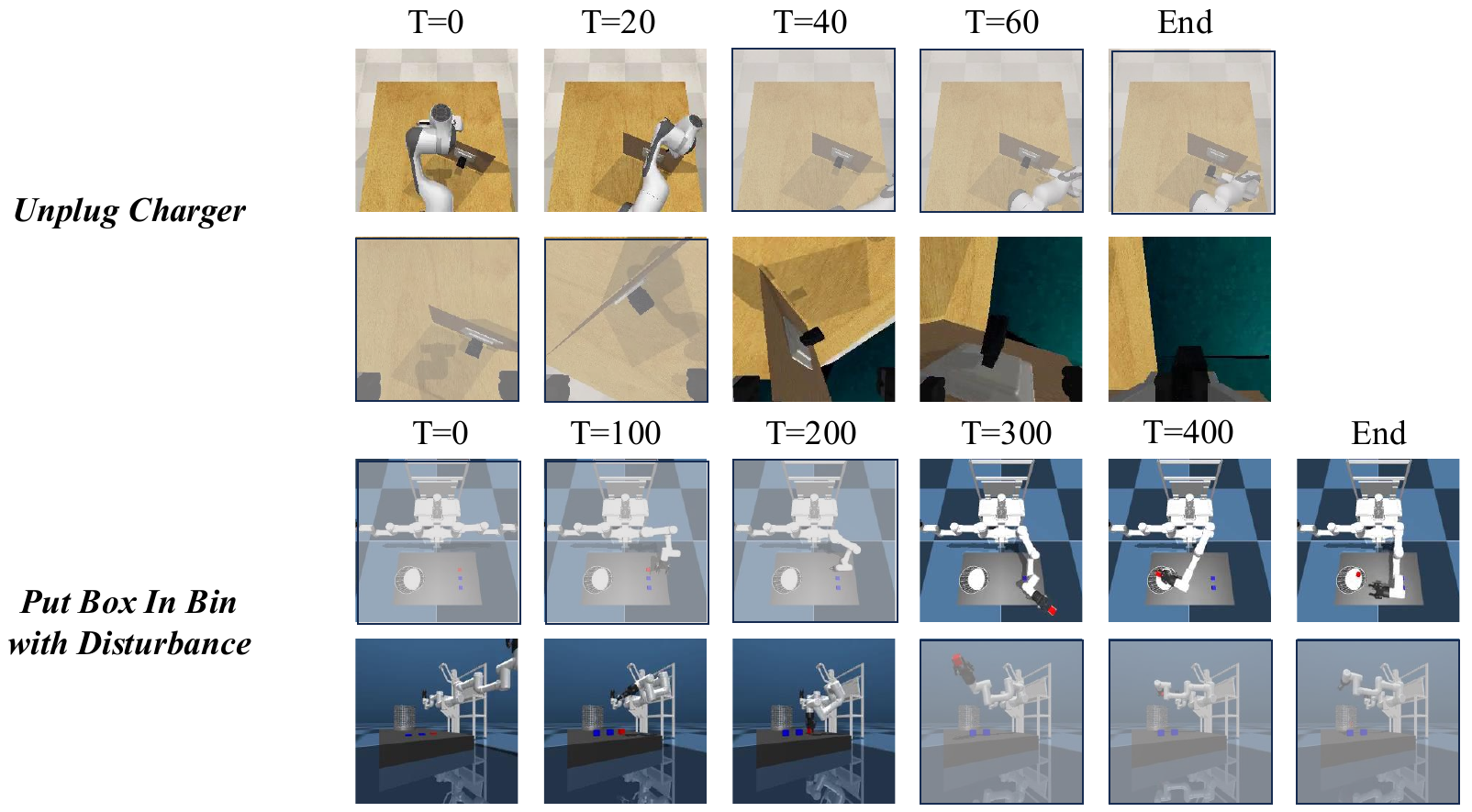}
\end{center}
   \caption{Visualization of the selected viewpoints in our simulation environments. The unmasked images represent the specific views selected by the model.}
\label{fig:vis_sim}
\end{figure}

\begin{figure}
\begin{center}
\includegraphics[width=\textwidth]{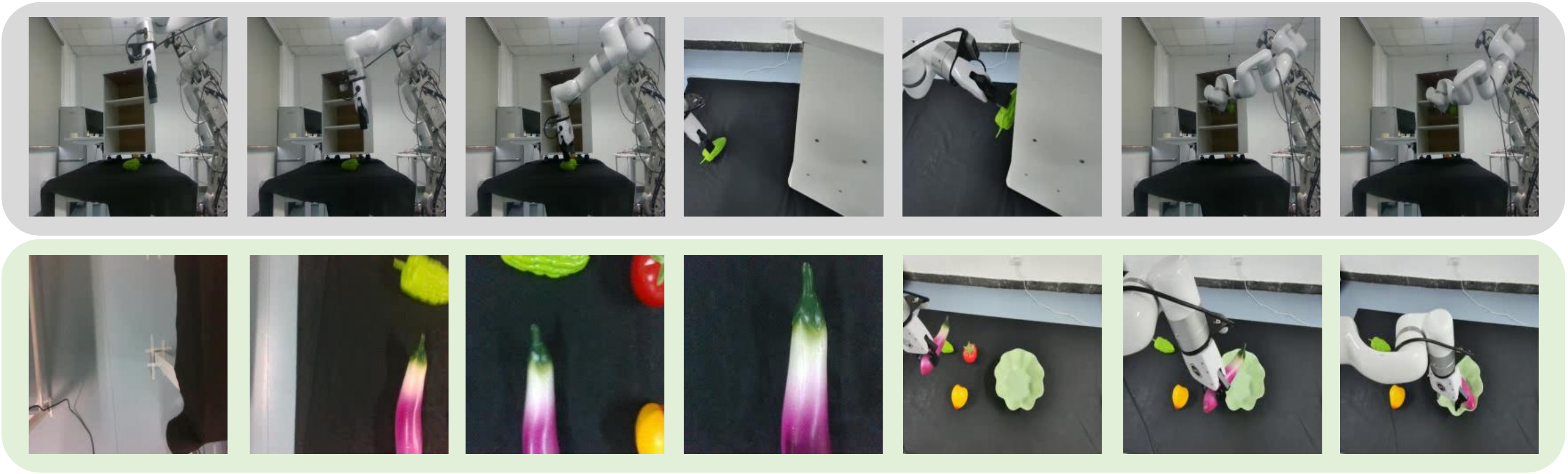}
\end{center}
   \caption{Visualization of the selected viewpoints in our real-world environments. Each row represents the procedure of a specific task, indicating the necessity of selecting different viewpoints.}
\label{fig:vis_real}
\end{figure}

\subsection{Visualization}

To intuitively understand the decision-making logic and behavioral patterns of the system under the single-camera control setup, we provide a comprehensive set of visualizations extracted during the deployment phases. As illustrated in Fig. \ref{fig:vis_sim} and Fig. \ref{fig:vis_real}, MAE-Select exhibits highly intelligent, human-attention-like decision-making capabilities—it dynamically and accurately selects the optimal viewpoint based on the task stage and contextual information. In the simulation environments (see Fig. \ref{fig:vis_sim}), we observe the temporal progression of the \textit{Unplug Charger} and \textit{Put Box In Bin with Disturbance} tasks. Taking \textit{Unplug Charger} as an example: in the early stages of the task (T=0 to T=20), the system selects the third-person viewpoint, a global perspective to establish the relative spatial arrangement of the robotic arm, charger, and socket. However, as the end-effector gradually approaches the target and requires high-precision alignment and extraction (T=40 and beyond), the system intelligently selects the wrist viewpoint. This dynamic transition effectively mitigates severe occlusions caused by the robot's own body, highlighting MAE-Select's ability to prioritize critical interaction areas while disregarding irrelevant background regions.

\section{Conclusions and Future Work}

In this work, we present MAE-Select, a novel framework that enables a robot to actively optimize its viewpoint for manipulation tasks using only a single camera. By fully leveraging pre-trained representations from multi-view masked autoencoders and dynamically selecting the next most informative viewpoints at each time chunk without manual annotations, MAE-Select significantly enhances the efficiency of robotic manipulation, addressing the limitations of both multi-camera and single-view setups. Our experiments demonstrate that this novel mechanism effectively improves performance, and even surpasses multi-camera systems in certain cases. Despite its effectiveness, one major limitation of MAE-Select is that it optimizes over discrete viewpoints rather than continuous ones, which reduces the system's flexibility in dynamic environments. Future improvements could involve the integration of techniques like Neural Radiance Fields (NeRF) \cite{mildenhall2021nerf} or 3D Gaussian processes \cite{kerbl20233d}, enabling continuous viewpoint optimization. 
\clearpage  


%
%
\bibliographystyle{splncs04}
\bibliography{main}
\end{document}